\documentclass{article}
\usepackage{authblk}

\usepackage{PRIMEarxiv}

\usepackage[utf8]{inputenc} 
\usepackage[T1]{fontenc}    
\usepackage{hyperref}       
\usepackage{url}            
\usepackage{booktabs}       
\usepackage{amsfonts}       
\usepackage{nicefrac}       
\usepackage{microtype}      
\usepackage{lipsum}
\usepackage{fancyhdr}       
\usepackage{graphicx}       
\usepackage{amsmath}
\usepackage{cite}
\graphicspath{{media/}}     

\title{A survey on fairness of large language models in e-commerce: Progress, application, and challenge}

\author[1]{Qingyang Ren}
\affil[1]{Department of Computer Science, Cornell Univerisity}

\author[2]{Zilin Jiang}
\affil[2]{Carnegie Mellon University}

\author[3]{Jinghan Cao}
\affil[3]{San Francisco State University}

\author[4]{Sijia Li}
\affil[4]{Carnegie Mellon University}

\author[5]{Chiqu Li}
\affil[5]{Columbia University}

\author[6]{Yiyang Liu}
\affil[6]{University of Southern California}

\author[7]{Shuning Huo}
\affil[7]{Virginia Tech}

\author[8]{Tiange He \thanks{Author contact Information: qr23@cornell.edu (Qingyang Ren), zilinjia@alumni.cmu.edu (Zilin Jiang), jcao3@alumni.sfsu.edu (Jinghan Cao), sijiali@alumni.cmu.edu  (Sijia Li), chiqu.l@columbia.edu (Chiqu Li), ianl@alumni.usc.edu (Yiyang Liu), shuni93@vt.edu (Shuning Huo), he.ti@northeastern.edu (Tiange He)}}
\affil[8]{Khoury College of Computer Sciences, Northeastern University}

\author[9]{Yuan Chen}
\affil[9]{New York University}

\begin{document}
\maketitle

\begin{abstract}
This survey explores the fairness of large language models (LLMs) in e-commerce, examining their progress, applications, and the challenges they face. LLMs have become pivotal in the e-commerce domain, offering innovative solutions and enhancing customer experiences. This work presents a comprehensive survey on the applications and challenges of LLMs in e-commerce.

 The paper begins by introducing the key principles underlying the use of LLMs in e-commerce, detailing the processes of pretraining, fine-tuning, and prompting that tailor these models to specific needs. It then explores the varied applications of LLMs in e-commerce, including product reviews, where they synthesize and analyze customer feedback; product recommendations, where they leverage consumer data to suggest relevant items; product information translation, enhancing global accessibility; and product question and answer sections, where they automate customer support.

The paper critically addresses the fairness challenges in e-commerce, highlighting how biases in training data and algorithms can lead to unfair outcomes, such as reinforcing stereotypes or discriminating against certain groups. These issues not only undermine consumer trust, but also raise ethical and legal concerns.

Finally, the work outlines future research directions, emphasizing the need for more equitable and transparent LLMs in e-commerce. It advocates for ongoing efforts to mitigate biases and improve the fairness of these systems, ensuring they serve diverse global markets effectively and ethically. Through this comprehensive analysis, the survey provides a holistic view of the current landscape of LLMs in e-commerce, offering insights into their potential and limitations, and guiding future endeavors in creating fairer and more inclusive e-commerce environments.
\end{abstract}


\section{Introduction}

The rapid advancement of LLMs has initiated a new era of natural language processing (NLP), revolutionizing various fields with their remarkable capabilities. Among these domains, e-commerce has emerged as a promising arena for the application of LLMs, offering innovative solutions and enhancing customer experiences. This survey investigates the fairness of LLMs in the e-commerce landscape, exploring their progress, applications, and the challenges they face.

The emergence of general LLMs, such as LLaMA \cite{touvron2023llama}, the GPT series \cite{openai2021gpt3, openai2024gpt4}, and Claude \cite{claude1, claude2}, has set new benchmarks in NLP tasks, including text classification, summarization, and question answering. Inspired by their remarkable success in general domains, the e-commerce sector has witnessed the rise of specialized LLMs tailored to its unique needs, such as understanding consumer behavior, optimizing search and recommendation systems, and automating content creation for product listings and marketing materials. Notable examples include EComGPT \cite{li2023ecomgpt} and E-BERT \cite{zhang2021ebert}, which have gained growing research interest in resolving pain points in both shopping and retailing experiences.

Despite the promising results achieved by existing e-commerce LLMs, several limitations and challenges remain to be addressed. These research gaps motivate the need for a comprehensive review that examines the fairness of LLMs in the e-commerce domain. There are concerns over existing e-commerce LLM regarding their potential to perpetuate harm. These models are typically trained on vast amounts of uncurated data sourced from the Internet, which can result in the inheritance of stereotypes, misrepresentations, derogatory language, and exclusionary behaviors. LLMs not only reflect existing biases but can also amplify them, leading to the automated perpetuation of injustice and the reinforcement of inequitable systems. The presence of biases in LLMs has been extensively documented, encompassing negative sentiment and toxicity towards specific social groups, stereotypical language associations, and a lack of recognition for certain language dialects \cite{hutchinson2020social, mechura-2022-taxonomy}.

\begin{figure}
    \centering
    \includegraphics[width=\linewidth]{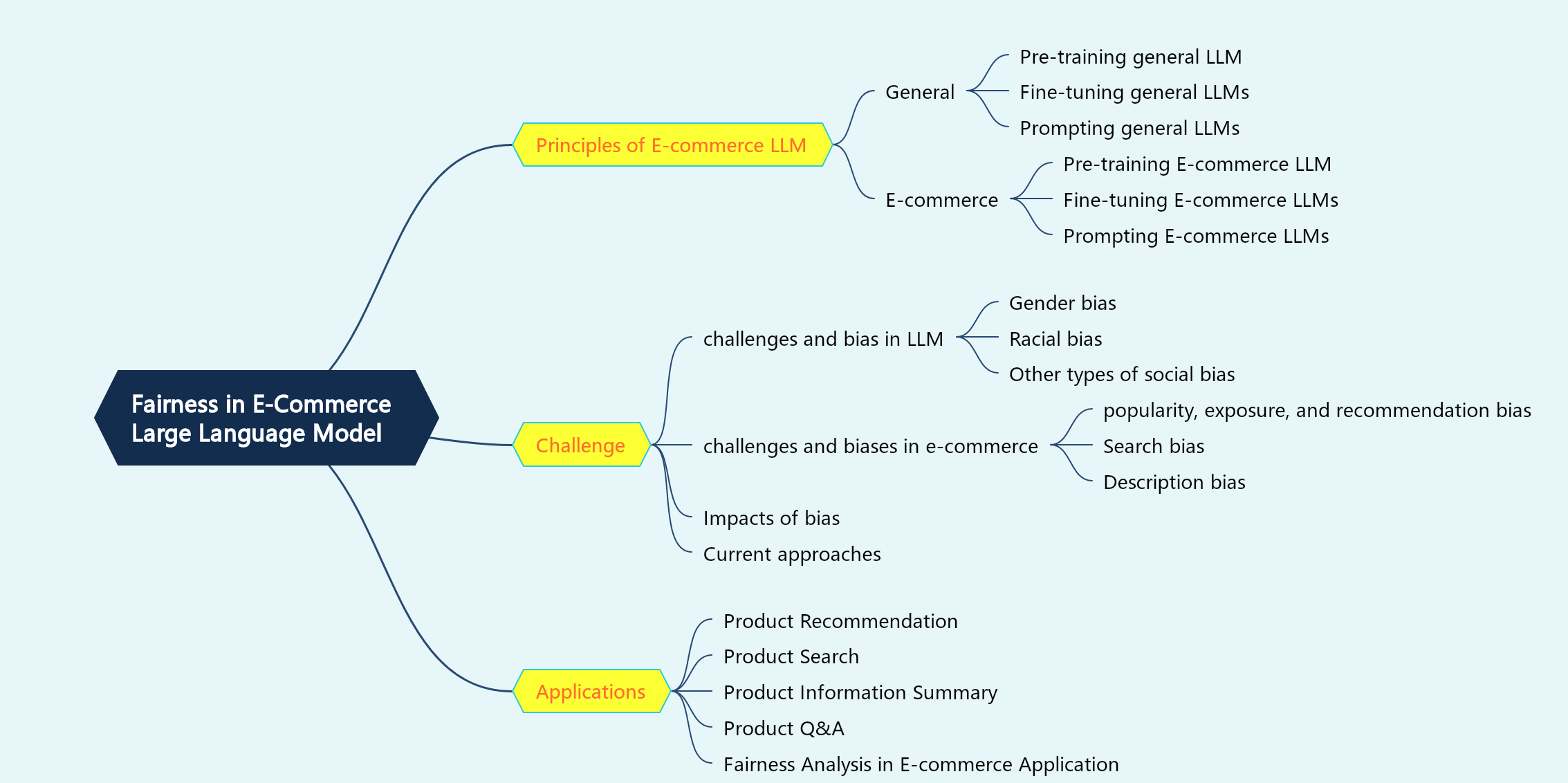}
    \caption{An overview of the fairness of e-commerce LLMs}
    \label{fig:enter-label}
\end{figure}

As shown in Figure \ref{fig:enter-label}, this survey is organized as follows: Section 2 summarizes the principles underlying the development of existing e-commerce LLMs, detailing the processes of pretraining, fine-tuning, and prompting that tailor these models to specific e-commerce needs. Section 3 describes the common fairness challenges faced by both general and e-commerce LLMs, shedding light on the potential biases and discriminatory outcomes that can arise from the training data and algorithms employed.

Section 4 outlines the outstanding applications of e-commerce LLMs, showcasing their versatility in areas such as product reviews, where they synthesize and analyze customer feedback; product recommendations, where they leverage consumer data to suggest relevant items; product information translation, enhancing global accessibility; and product question and answer sections, where they automate customer support. However, this section also critically examines the specific fairness challenges that arise within each application domain, highlighting the potential for biases and unfair outcomes that can undermine consumer trust and raise ethical and legal concerns.

Finally, Section 5 explores future research directions, emphasizing the need for more equitable and transparent LLMs in e-commerce. It advocates for ongoing efforts to mitigate biases and improve the fairness of these systems, ensuring they serve diverse global markets effectively and ethically. Through rigorous evaluation and continuous improvement, e-commerce LLMs can foster inclusive and trustworthy online shopping experiences, benefiting both consumers and businesses alike.
Through this comprehensive analysis, the survey provides a holistic view of the current landscape of LLMs in e-commerce, offering insights into their potential and limitations, and guiding future endeavors in creating fairer and more inclusive e-commerce environments. By addressing the fairness challenges directly and promoting responsible development and deployment of LLMs, the e-commerce sector can harness the full potential of these powerful models while upholding ethical principles and safeguarding consumer rights.

\section{The Principles of E-commerce LLMs}
LLM training comprises three major different approaches: pre-training, fine-tuning and prompting. Given the system complexity in E-commerce domain, the relevant research is shifting from applying a single model training to an integration of multiple LLM model, of which is trained for specific tasks in a larger system.

\begin{table}[ht!] 
 \caption{Summary of existing LLMs in E-commerce}
  \centering \resizebox{\columnwidth}{!}{
  \begin{tabular}{llllll}
    \toprule
    Domain     & Model Type  & Model  & Base & Param  & Data Source \\
    \midrule
  General    & Pre-training  & ALBERT\cite{lan2019albert}           & BERT & 12M/18M/60M/235M & BooksCorpus, English Wikipedia \\
    General    & Pre-training  & BERT\cite{devlin2019bert}            & -    & 110M/340M       & BooksCorpus, English Wikipedia\cite{devlin2019bert} \\
    General    & Pre-training  & BART \cite{lewis2020bart}            & -    & 140M/400M       & mix of books and Wikipedia data \\
    General    & Pre-training  & ELECTRA\cite{clark2020electra}       & -    & 14M/110M/335M   & BooksCorpus, English Wikipedia \\
    General    & Pre-training  & XLNet\cite{yang2019xlnet}            & -    & 110M/340M       & Wikipedia, BookCorpus \\
    General    & Pre-training  & ERNIE\cite{sun2019ernie}             & -    & 110M            & Wikipedia, other texts \\
    General    & Pre-training  & Galactica\cite{galactica2022}        & -    & 6.7B/30.0B/120.0B & Scientific papers \\
    General    & Pre-training  & GPT-2\cite{radford2019language}      & -    & 1.5B            & WebText \\
    General    & Pre-training  & DeBERTa\cite{he2021deberta}          & BERT & 1.5B            & BooksCorpus, English Wikipedia \\
    General    & Pre-training  & LLaMA\cite{llama2023}                & -    & 7B/13B/33B/65B  & Diverse internet data \\
    General    & Pre-training  & LLaMA-2\cite{touvron2023llama}       & -    & 7B/13B/34B/70B  & Larger dataset \\
    General    & Pre-training  & GPT-3\cite{brown2020language}        & -    & 6.7B/13B/175B   & Extensive internet text \\
    General    & Pre-training  & PaLM\cite{chowdhery2022palm}                                 & -    & 8B/62B/540B     & Public and proprietary data \\
    General    & Fine-tuning   & Alpaca\cite{stanford_alpaca}         & LLaMA & 7B/13B          & LLaMA datasets, additional data \\
    General    & Fine-tuning   & InstructGPT \cite{ouyang2022training}& -    & 175B            & Based on GPT-3 \\
    General    & Fine-tuning   & GPT-4 \cite{openai2024gpt4}          & -    & -               & - \\
    General    & Fine-tuning   & Mixtral \cite{jiang2024mixtral}      & -    & 8x7B            & multilingual data using a context size of 32k tokens \\
   
    E-commerce & Pre-training    & E-BERT  \cite{zhang2021ebert}    & BERT & 110M    & Amazon Dataset    \\ 
    E-commerce & Pre-training    & KG-FLIP \cite{Jia2023}     & BLIP & 224M    & Amazon Dataset    \\ 
    E-commerce & Fine-tuning  & Ecom-GPT\cite{li2023ecomgpt}  & BLOOMZ & 560M & EcomInstruct     \\
    E-commerce & Fine-tuning      & G2ST\cite{chen2024general2specialized}  &Qwen-14B  & 14B & Alibaba.com    \\ 
    E-commerce & Fine-tuning    & eCeLLM-L\cite{peng2024ecellm}     & Flan-T5-XXL, Llama-2-13B-chat & 11B-13B    & ECInstruct \cite{peng2024ecellm}  \\ 
    E-commerce & Fine-tuning   & eCeLLM-M\cite{peng2024ecellm}     & Llama-2-7B-chat, Mistral-7B & 7B    & ECInstruct \cite{peng2024ecellm}    \\ 
    E-commerce & Fine-tuning   & eCeLLM-S\cite{peng2024ecellm}     & Flan-T5-XL-3B, Phi-2-3B & 3B    & ECInstruct \cite{peng2024ecellm}   \\ 
    E-commerce & Fine-tuning    & GPT4Rec\cite{li2023gpt4rec}     & GPT-2    & 117M & Amazon Review: Beauty and Electronics \cite{li2023gpt4rec}   \\ 
    E-commerce & Prompt-tuning    & MixPAVE\cite{yang-etal-2023-mixpave}  & Pre-training Transformer \cite{pre-trained-transformer-mave} & 0.445M & AE-110K \cite{xu-etal-2019-scaling} , MAVE\cite{yang-etal-2023-mixpave}    \\
    E-commerce & Prompt-tuning    & CTM\cite{wang2022continuous}  & characterBERT , BERT & - & Huski.ai   \\
    E-commerce & Prompt-tuning    & Aspect Extraction LLM\cite{li2023prompt}  &GPT-2,BERT  & - & Amazon, Yelp, Tripadvisor  \\
    E-commerce & Prompt-tuning    & CF Recommender Enhancement Model\cite{dang2022enhancing}  & BERT,RoBERTa   & - &  Amazon US Reviews  \\
    E-commerce & Prompt-tuning    & recGPT\cite{dang2022enhancing}  & pre-trained ChatGPT   & - &  Amazon reviews and Yelp  \\ 
    \bottomrule
  \end{tabular}}
  \label{tab:table}
\end{table}
\subsection{Pre-training}

\begin{figure}[ht!]
\centering
\includegraphics[width=160mm]{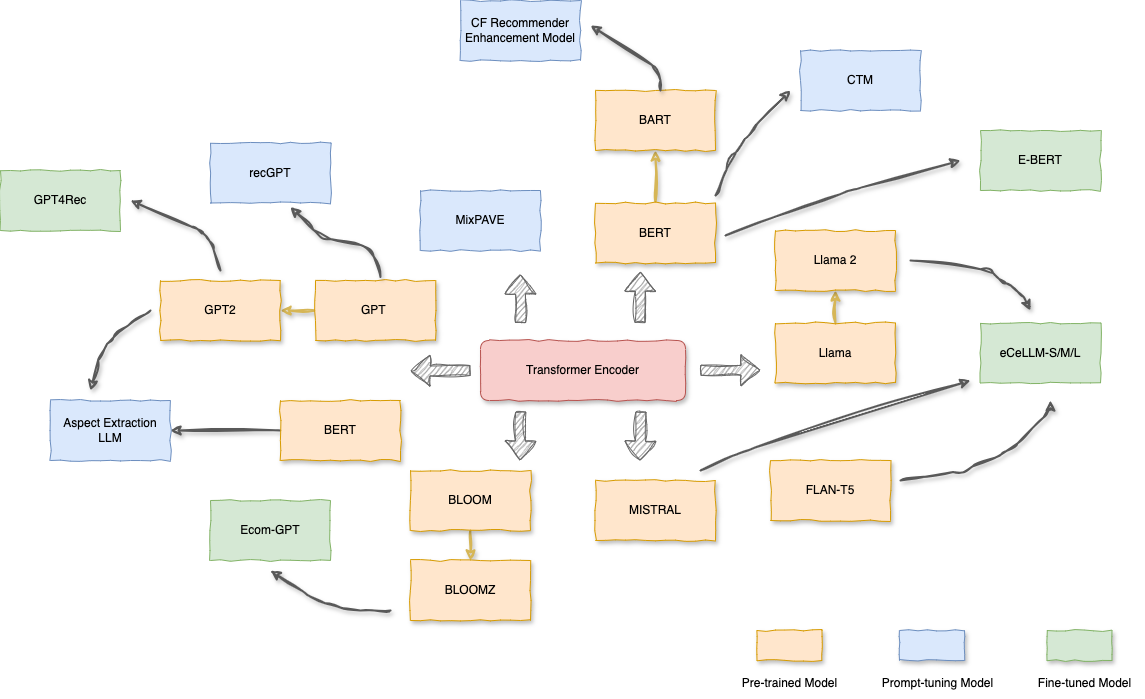}
\caption{LLM Dependency in E-commerce 
}
\end{figure}

Pre-training involves training a large language model (LLM) from scratch on a substantial corpus of e-commerce texts. This foundational training equips the model with domain-specific knowledge necessary to tackle a wide range of tasks within the e-commerce sector. Utilizing foundational architectures such as the Transformer \cite{vaswani2017attention} and subsequent adaptations like BERT \cite{devlin2019bert} and GPT \cite{openai2021gpt3}, these models are pre-trained on extensive corpora, including product descriptions, customer reviews, and user interactions. These datasets enable the models to capture a variety of linguistic nuances and e-commerce specific terminologies. For instance, BERT's bidirectional training structure is particularly effective for tasks like sentiment analysis and intent recognition, which are crucial for personalized product recommendations and customer service automation \cite{devlin2019bert} . Similarly, GPT's autoregressive features are adept at generating coherent and engaging product descriptions that can significantly enhance search engine optimization (SEO) and user interaction \cite{openai2021gpt3}.

In e-commerce field, specialized models such as E-BERT \cite{zhang2021ebert} and KG-FLIP \cite{Jia2023} further refine these capabilities. E-BERT, a derivative of BERT, is re-pre-trained on the Amazon Dataset to boost its efficacy in product-related tasks, thereby enhancing customer interaction quality and the accuracy of sentiment analysis. On the other hand, KG-FLIP extends this by integrating knowledge graphs, which enrich the model's understanding of structured product information and customer data, leading to improved contextual awareness and precision in search functionalities.

Future research is likely to focus on refining pre-training approaches to better handle the informal and varied nature of e-commerce text, expand multilingual support, and enhance context-aware systems, potentially incorporating newer models such as GPT-4, LLaMA-3 for even more robust applications.

\subsection{Fine-tuning}
There's a lack of consensus on the precise definition of "fine-tuning" (also known as model tuning \cite{devlin2019bert}) within the industry as it emerges with the iteration of researchers experimenting on pre-trained models. Fine-tuning is based on an existing model and then further trained with specific datasets of samples and parameter-efficient tuning approaches such as Lora \cite{hu2022lora}, Prefix-tuning \cite{li-liang-2021-prefix} and full parameter tuning \cite{lv2023parameter}. In E-commerce, precedent researches have emphasized the feasibility of applying fine-tuned LLMs to address specific tasks. Li et al.\cite{li2023ecomgpt} (2023) proposed a Ecom-GPT model which was trained based on BLOOMZ\cite{muennighoff2023crosslingual} with instruct datasets. In zero-shot scenarios, this model shows outperformed metrics than other general LLMs in terms of attribution extraction, customer topic classification and product title generation. Chen et al. \cite{chen2024general2specialized}(2024), targeting the translation tasks in E-commerce, offered a general-to-specialized paradigm based on Neural Machine Translation models. Two-step fine-tuning approaches are incorporated into the experiment with 14 billion parameters trained on bilingual datasets. The ROUGE-N and ROUGE-L metrics reveals a better results in translation tasks compared with LLaMA, Qwen and GTP. Peng et al. \cite{peng2024ecellm} (2024) has proposed a set of E-commerce models (eCeLLM) to strengthen the generalization abilities including product understanding, user request understanding, product matching and question answering. Particularly, three different model sizes are developed and compared with general-purpose LLMs, e-commerce LLMs and task-specific models given a comprehensive set of individual tasks. It is noted that the models demonstrate higher F1 scores and better generalization ability in out-of-domain test cases. Li et al.\cite{li2023gpt4rec} (2023) fine-tuned GPT-2 with a 2-step training process and then integrate it with a search engine. This framework aims to leverage LLM to generate recommended products given the customers previous purchase history.    

Apart from relevant works on improving single model's performance for either specific tasks or generalization ability in E-commerce, researchers also proposed novel systematic integration of multiple trained LLMs to handle customer requests in real-world applications. Zhou et al.\cite{zhou2023enhanced} (2023) proposed an approach to synthesize fine-tuned BERT and Llama 2 in a system to efficiently extract product attributes for customer queries in Walmart search functionality. When customers query for specific products, the most likely matched results will be returned. This system employs BERT to generate contextual embedding as the encoding phase and Conditional Random Field(CRFs) \cite{10.3115/1219840.1219885} layer to decode the tags. In parallel the encoding from BERT is utilized and trained to construct neural network providing decorative relation correction scrutinizing on the returned responses. This system not only incorporate a fine-tuned BERT as base model but also leverage LLAMA 2.0 with prompts to retrieve additional product attributes for customers. Another practice proposed by Zhao et al.\cite{zhao2024breaking} (2024) also utilizes BERT-CRF model in encoding/decoding for entity extraction. The difference is that Zhao et al.\cite{zhao2024breaking} (2024) builds a complementary graph (Entity Dict) to recommend the next products for customers. Cloude 2, as the pre-trained model, is fine-tuned to discern the complementary relationships in the graph construction. 

\subsection{Prompt-tuning}
\begin{figure}[ht!]
\centering
\includegraphics[width=90mm]{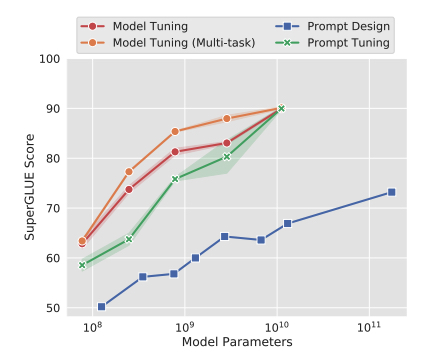}
\caption{Prompt tuning of T5 performance as size increases \label{overflow}
\label{fig:prompting}
\cite{lester2021power}}
\end{figure}

Prompt-tuning, as Lester et al.\cite{lester2021power} explained, is a  mechanism of freezing language models and tuning model with task-specific prompt for each task. The whole process is effective in terms of only a few tun-able tokens prepend per downstream
task and grouping multiple adaptations of a pre-trained language model to achieve similar or even better performance than traditional fine-tuning approach (See Figure~\ref{fig:prompting}). 

Lester et al.\cite{lester2021power}'s work reveals the lightweight parameter footprint and multi-task serving in prompt-tuning. Based on such findings, there have been growing interests in applying prompt-tuning in E-commerce. Yang et al.\cite{yang-etal-2023-mixpave} (2023) proposed a mix-prompt trained model for product attribute value extraction(MixPAVE). The base pre-trained transformer model is frozen except the extraction head\cite{NIPS2014_375c7134} to be trained with textual and key-value prompts. Additionally, two datasets: MAVE\cite{mave-2021} with 3 million attribute-value annotations and AE-110K collected from AliExpress are used for training and partially for testing in few-shot scenarios. The experiment result shows that MixPAVE outperforms fine-tuning models regarding certain attributes extraction with fewer parameters trained over the process. In a similar study on prompt-tuning, Wang et al.\cite{wang2022continuous} observed a significant rise in the number of new entities emerging in the E-commerce domain. To overcome limitations of existing LLM's ability to handle emerging product entities and titles, Wang et al.\cite{wang2022continuous} proposes a textual entailment model with continuous prompt-tuning approach to better classify entity types. The experiment result shows higher average F1 score in both addition (0.30\%) and concatenation (0.38\%) as fusion methods.  

In some research, prompt-tuning is jointly applied with fine-tuning approach. For instance, Li et al.\cite{li2023prompt} (2023) fine-tuned GPT-2 with local offline datasets and then feed the model with soft prompts concatenated with embeddings from customer review texts to generate a list of aspect terms. These terms successively are fed into a neural network to generate aspect-based recommendations. With a chain of fine-tuned LLMs and prompt-tuning method connected to aspect-based recommender systems, the frameworks shows better metrics than state-of-the-art baseline methods in providing more meaningful recommendations for users. 

Moreover, some researcher intentionally blurs the boundary of prompt-tuning and fine-tuning  to serve specific purpose. For example, Dang et al.\cite{dang2022enhancing} fine-tuned BERT and RoBERTa with prompt-based learning paradigm to generate more sentiment data in order to tackle the insufficient rating data and data sparsity issues in collaborative filtering recommendation systems. 



\section{Bias challenges}
\label{sec:others}

\subsection{Introduction to Fairness and Bias in LLM, E-commerce}

LLMs and AI systems have revolutionized the e-commerce industry, enabling personalized experiences and efficient decision-making processes. However, these advanced technologies also introduce challenges related to fairness and bias. LLMs can perpetuate and amplify societal biases present in their training data, leading to discriminatory outcomes and unfair treatment of certain demographics. In the e-commerce domain, biases can manifest in various forms, such as popularity bias in product recommendations, exposure imbalance among sellers, and skewed search results. These biases not only impact user experiences but also raise ethical concerns regarding transparency, accountability, and equity. Addressing fairness and bias in LLMs and e-commerce requires a multifaceted approach, including the development of fairness-aware algorithms, diverse and representative training data, and rigorous evaluation frameworks. Ongoing research efforts and interdisciplinary collaboration are crucial to mitigate biases, ensure fair outcomes, and build trust in AI-driven e-commerce systems. As the integration of LLMs and AI continues to shape the future of e-commerce, prioritizing fairness and addressing bias remains paramount for creating inclusive and equitable online marketplaces.

\subsection{What is fairness and bias?}

Fairness and bias are two interrelated concepts that are crucial in the context of AI and machine learning systems. Fairness refers to the principle of ensuring equitable treatment and outcomes for all individuals or groups, regardless of their protected attributes such as race, gender, age, or socioeconomic status. It involves the absence of discrimination or unjustified disparities in the decisions or outputs generated by AI algorithms. On the other hand, bias refers to the systematic errors or prejudices that can be present in data, algorithms, or models, leading to skewed or unfair results. Bias can arise from various sources, including biased training data, flawed data collection processes, or the inherent limitations of the algorithms themselves. Biases can manifest in different forms, such as demographic biases, measurement biases, or representation biases, and can perpetuate or amplify existing societal inequalities. Ensuring fairness and mitigating bias in AI systems is essential to prevent discriminatory outcomes, promote equal opportunities, and build trust in the technology. It requires a proactive approach that involves careful data curation, algorithmic fairness techniques, rigorous testing and evaluation, and ongoing monitoring to identify and address any potential biases throughout the AI lifecycle.

\subsection{The challenges and biases in LLM}

LLMs have made remarkable advancements in natural language processing, but they also face significant challenges related to fairness and bias. One major challenge is the presence of various types of biases in LLMs, including gender bias, racial bias, religious bias, age bias, sexuality bias, country bias, and disease bias. These biases can manifest in the model's outputs, leading to stereotypical or discriminatory associations. Another challenge lies in the sources of bias in LLMs, which can stem from biased training data, sampling biases, semantic biases encoded in the model's representations, and the amplification of biases during the learning process. Addressing these challenges requires a comprehensive approach that involves careful data curation, bias mitigation techniques, and rigorous evaluation frameworks. Additionally, the development of explainable and interpretable LLMs is crucial to understand and mitigate biases effectively. Researchers and practitioners must also consider the ethical implications of deploying LLMs in real-world applications and ensure that the models align with principles of fairness, transparency, and accountability. Overcoming these challenges is essential to harness the full potential of LLMs while promoting fairness and reducing the risk of perpetuating societal biases.

\subsubsection{Gender bias}

LLMs can exhibit various types of biases that pose significant challenges to their fairness and reliability. Among them, gender bias \cite{Kotek_2023} is a prominent issue, where LLMs may associate certain occupations or attributes with specific genders, perpetuating stereotypical assumptions. Gender bias has been demonstrated to be present in word embeddings, as well as in a wide range of models designed for diverse NLP tasks, including machine translation, sentiment analysis, auto-captioning, toxicity detection, and beyond. Since LLMs often failed to acknowledge the ambiguity in pronoun references unless explicitly prompted, LLMs often provided explanations that appeared logical but were factually inaccurate, potentially masking the biases.
One significant source that introduces gender bias is labelling \cite{li2024survey}, which occurs when the training data contains biased or subjective labels provided by annotators, leading the model to learn and perpetuate those biases. If the training data for sentiment analysis predominantly associates certain genders with specific sentiments, such as associating women with emotions like "sensitivity" and men with "strength", an LLM might learn and reinforce these stereotypes. For instance, it may consistently associate pronouns referring to women with negative sentiments or pronouns referring to men with positive ones. 

One potential solution to mitigate gender bias in LLM is to ensure that training datasets are diverse and representative of different genders, races, cultures, and backgrounds. This involves collecting data from a wide range of sources and demographics to minimize biases present in the data.

\subsubsection{Racial bias}

Racial bias \cite{yang2024unmasking} can also be present, leading to biased outputs or decisions based on race-related information. The models tended to generate biased content for certain racial groups, including unwarranted details based on race. The models exhibited favoritism and has racially-skewed socio-economic projection towards a certain racial group in content recommendations. A important origin of this racial bias is sampling. The issue arises when the distribution of samples from different demographic groups in the training data differs from the actual population distribution, causing the model to exhibit biased behavior. Pre-existing racial prejudices and inequalities within the data can be reflected in the outputs of the language models. Additionally, the vulnerability of the models to prompt manipulation with malicious intent can lead to biased responses.

One potential solution to mitigate racial bias in LLM is to analyze the distribution of racial groups within the data and adjusting the sampling process to ensure equal representation. Random sampling techniques could be adopted to select data for training the language models, which helps reduce the risk of bias by ensuring that each data point has an equal chance of being selected, regardless of racial characteristics. Stratified sampling can be employed to ensure proportional representation of different racial groups in the training data, which involves dividing the dataset into strata based on race and then sampling proportionally from each stratum to ensure balanced representation.

\subsubsection{Other types of social bias}

Religious bias occurs when LLM demonstrates favoritism or discrimination towards individuals or groups based on their religious beliefs or affiliations. It may originate from the generation of text that stereotypes or stigmatizes certain religions, promotes one religion over others, or misrepresents religious practices and beliefs. Similarly, LLM could cause and even amplify other social bias including age bias, sexuality bias, and country bias. Potential source of these social bias could be semantic bias, which can emerge during the language model encoding process, resulting in biased semantic representations that capture stereotypical associations. These social bias could also be amplified, where the model not only learns the biases present in the training data but also amplifies them during the learning process. They can persist and even intensify further when the model is fine-tuned for downstream tasks. 
 
To mitigate the bias in LLMs requires careful attention to data quality and representative sampling during both pre-training and fine-tuning stages. It also involves developing robust evaluation frameworks to detect and quantify biases, enabling researchers to identify and address them effectively. By understanding and tackling the sources of bias, we can work towards building more fair and unbiased LLMs that provide reliable and equitable outputs.

\subsection{What are the challenges and biases in e-commerce?}
E-commerce platforms face several challenges and biases that can impact the fairness and equity of the online marketplace. One significant challenge is the presence of various types of biases, such as popularity bias, where popular items or sellers receive disproportionate exposure in recommendation systems, hindering the visibility of less popular offerings. Exposure bias refers to the skewed distribution of visibility and opportunities among sellers, with a small percentage of popular sellers receiving the majority of user attention. This can lead to unfair competition and limit the growth potential of smaller or newer sellers. Recommendation bias can also arise, where the algorithms used to suggest products to users are influenced by factors beyond relevance or user preferences, leading to the promotion of certain products or sellers over others. Search bias can further compound these issues, as the search results on e-commerce platforms may be skewed towards certain products or sellers due to factors such as search optimization techniques or paid placements.Moreover, e-commerce platforms must grapple with description bias, where the textual metadata like product tags and descriptions provided by sellers may not accurately or comprehensively reflect the offerings. Addressing these challenges and biases, which also include ensuring seller-side fairness and providing fair and unbiased product reviews, requires the implementation of fair and transparent algorithms, robust evaluation frameworks, and a commitment to creating an equitable online marketplace. By promoting fairness and mitigating these varied biases, e-commerce platforms can build a more trustworthy and inclusive digital environment that benefits both sellers and consumers.

\subsubsection{Visibility and accessibility: popularity, exposure, and recommendation bias}
Visibility and accessibility are crucial to products on e-commerce platforms. One prominent issue is popularity bias, where popular items or sellers receive disproportionate exposure and visibility in recommendation systems, overshadowing less popular offerings. This bias can limit the discoverability of new or niche products and hinder the growth of smaller sellers \cite{klimashevskaia2023survey}. Another type of bias is exposure bias, which refers to the skewed distribution of visibility and opportunities among sellers, with a small group of popular sellers receiving the majority of user attention and sales. Less exposed items pose the challenge of inaccurate reward function prediction in our e-commerce setting \cite{sellersidefairness}. This bias can create an uneven playing field and stifle competition. As discussed in prior research, the concept of higher-ranked items in recommendation lists commonly receiving more exposure and user attention, and being more likely to be consumed, was also addressed \cite{joachims2007evaluating, abdollahpouri2020addressing}.

Recommendation bias occurs when the algorithms used to suggest products to users are influenced by factors beyond relevance or user preferences, such as promotional partnerships or business objectives \cite{klimashevskaia2023survey}. This bias can lead to the promotion of certain products or sellers over others, potentially compromising the integrity of the recommendations and creating an uneven playing field \cite{klimashevskaia2023survey, joachims2007evaluating, abdollahpouri2020addressing}.

To address recommendation and other biases related to visibility and accessibility, recent research has proposed bias mitigation strategies that go beyond relying solely on binary rating matrices \cite{cikm2021}. These more advanced techniques require complex model adjustments, expensive sampling methods, or heuristic propensity scores, and can struggle when users accept or reject multiple recommendations for the same item \cite{cikm2021}. An alternative approach, as suggested in the literature, is a multi-process fusion method that combines pre-processing, in-processing, and post-processing techniques to alleviate popularity bias in recommendations \cite{tece2023}. This approach embeds consumer preferences and product popularity information directly into the recommendation model, while also making adjustments to the dataset and recommendation lists, without imposing specific requirements on the underlying recommendation algorithm \cite{tece2023}. This multi-faceted debiasing strategy has been shown to improve recommendation accuracy and consumer interest, making it a promising solution for addressing recommendation bias in e-commerce and LLM-powered systems.  

\subsubsection{Search bias}
Platforms facilitating digital commerce have long grappled with the challenge of search bias, where search results are skewed towards certain products or sellers due to factors such as search optimization techniques or paid placements. This bias in search can adversely impact the visibility and discoverability of products, thereby affecting consumer choice and fair competition. Addressing such biases necessitates the implementation of fair and transparent search algorithms, regular audits and evaluations, and a steadfast commitment to fostering an equitable e-commerce ecosystem for all participants.

To combat the issue of search bias, a novel model training framework dubbed "TripleLearn" was proposed \cite{cheng2023end}. The cornerstone of this solution is that TripleLearn iteratively learns from three distinct training datasets, deviating from the traditional approach of employing a single training set. By harnessing this iterative learning process, the authors were able to substantially enhance the model's performance, boosting the F1 score from 69.5 to an impressive 93.3 on the holdout test data \cite{cheng2023end}. This remarkable improvement underscores the efficacy of the TripleLearn approach in mitigating search bias and delivering high-quality search results for e-commerce platforms, thereby fostering a more equitable and transparent search experience.

\subsubsection{Description bias}
E-commerce platforms have long been grappled with the challenge of description bias, which stems from the manner in which sellers provide textual metadata, such as product tags, to characterize their offerings. As these digital marketplaces facilitate active user participation in the creation and categorization of product-related content, the textual features (e.g., titles, descriptions, tags) generated by sellers may not always be of sufficient quality or accurately reflect the nuances of the products \cite{belem2023fixing}. Sellers, lacking the comprehensive training or domain expertise required to meticulously describe their wares, may instead resort to the use of "tag spam" -- the employment of irrelevant yet popular keywords in a misguided attempt to promote their products \cite{belem2023fixing}. This seller-generated bias in product descriptions can have detrimental impacts on the efficacy of crucial e-commerce services, such as search and recommendation systems, ultimately frustrating consumers' ability to effectively locate their desired items \cite{belem2023fixing}. Addressing this description bias is of paramount importance for enhancing the quality of textual product descriptors and, correspondingly, improving the overall e-commerce user experience.

In an effort to alleviate the description bias introduced by seller-provided product tags, the scholarly work under consideration proposes the leveraging of automated tag recommendation techniques that harness search query and click data \cite{belem2023fixing}. The central hypothesis posits that "the set of queries collectively issued by the consumers of the e-marketplace, along with corresponding clicks, reflect a more trustworthy view of the products; thus those queries and clicks can be exploited as a source of high-quality (e.g., more diverse) tags to describe the products" \cite{belem2023fixing}. Guided by this principle, the authors develop novel tag recommendation solutions, including deep learning-based approaches, that generate tags based on the insights gleaned from this search data \cite{belem2023fixing}. Rigorous evaluations revealed that the seller-provided tags often contain significant noise and bias, while the proposed search-boosted tag recommenders were able to substantially outperform the state-of-the-art, improving recommendation effectiveness by over 16 per cent \cite{belem2023fixing}. The authors contend that these recommended tags, borne of the collective consumer search experience, can provide a more reliable data source for e-commerce search and information services than the original seller-provided descriptions, thereby helping to overcome the inherent biases \cite{belem2023fixing}.

\subsection{What are the challenges and biases in the application of LLM in the e-commerce field?}
The application of LLMs in the e-commerce field presents several challenges and biases that need to be addressed to ensure fairness and equity. One significant challenge is the potential for LLMs to perpetuate and amplify biases present in the training data, which can lead to discriminatory outcomes in e-commerce recommendations, search results, and customer interactions. For example, if an LLM is trained on biased product descriptions or customer reviews, it may generate biased outputs that favor certain products or sellers over others. One example is the Modern collaborative filtering algorithms seek to provide personalized product recommendations by uncovering patterns in consumer product interactions: Addressing Marketing Bias in Product Recommendations \cite{addressingmarketingbias}. Additionally, LLMs may struggle to capture the nuances and context-specific meanings of e-commerce terminology, leading to misinterpretations or inaccurate recommendations. Another challenge is the lack of transparency and explainability in LLM-based e-commerce systems, making it difficult to identify and mitigate biases. Moreover, the application of LLMs in e-commerce may raise privacy concerns, as these models require vast amounts of user data for training and operation. Ensuring the responsible and ethical use of user data while maintaining the benefits of personalization is a delicate balance. To address these challenges, e-commerce platforms must prioritize the development of fair and unbiased LLMs, incorporate diversity and inclusivity in training data, and implement robust evaluation and auditing mechanisms. Collaboration between e-commerce practitioners, researchers, and ethicists is crucial to navigate the ethical implications and ensure the responsible deployment of LLMs in the e-commerce field \cite{addressingmarketingbias}.

\subsection{Impacts of bias}
Bias in AI systems and e-commerce platforms can have far-reaching and detrimental impacts on individuals, businesses, and society as a whole. One significant impact is the perpetuation and amplification of societal inequalities and discrimination. Biased algorithms can lead to unfair treatment of certain demographics, limiting their access to opportunities, resources, and services. This can result in a widening of the digital divide and the reinforcement of historical biases \cite{addressingmarketingbias}. Moreover, biased AI systems in e-commerce can lead to discriminatory outcomes, such as skewed product recommendations, unfair pricing, or biased search results. This can harm the reputation and trust in e-commerce platforms, as consumers may feel misled or unfairly treated. Bias can also have economic consequences, stifling competition and innovation by favoring established or popular brands over newer or niche offerings. This can create barriers to entry for small businesses and limit consumer choice \cite{addressingmarketingbias}. Additionally, biased AI systems can perpetuate stereotypes and contribute to the spread of misinformation, influencing public opinion and decision-making. The impacts of bias extend beyond the individual level, affecting society's collective values, beliefs, and behaviors. Addressing the impacts of bias requires a proactive and multifaceted approach, including the development of fair and transparent AI systems, regular audits and assessments, and the promotion of diversity and inclusivity in the design and deployment of AI technologies.

\subsection{Current approaches}
Addressing the challenges of fairness and bias in LLMs and e-commerce platforms requires a multifaceted approach. Current efforts focus on developing fairness-aware algorithms that incorporate fairness metrics and constraints into the training and evaluation processes \cite{dash2022alexa}. These algorithms aim to mitigate biases by ensuring equitable treatment of different groups and promoting diversity in the model's outputs. Another approach is the use of adversarial debiasing techniques, which involve training the model to be invariant to sensitive attributes, such as gender or race, while still maintaining its predictive performance \cite{dash2022alexa}. Researchers are also exploring the use of counterfactual fairness frameworks, which assess the fairness of a model by considering the outcomes under different hypothetical scenarios. In the e-commerce domain, current approaches include the development of fair ranking algorithms that ensure equitable exposure for all sellers and products, regardless of their popularity or historical performance. Collaborative filtering techniques are being adapted to incorporate fairness constraints and promote diversity in recommendations. Additionally, there is a growing emphasis on transparency and explainability in e-commerce algorithms, allowing stakeholders to understand and audit the decision-making processes \cite{dash2022alexa}. Efforts are also being made to curate diverse and representative training data to reduce the impact of historical biases. Overall, the current approaches to tackling fairness and bias in LLMs and e-commerce involve a combination of algorithmic innovations, data curation strategies, and transparency initiatives to ensure equitable outcomes for all participants in the digital marketplace \cite{dash2022alexa}.

\section{E-commerce Application}
The integration of language models such as ChatGPT has transformed the customer-business interaction within e-commerce applications. By harnessing the extensive knowledge and linguistic capabilities of these models, e-commerce platforms can deliver personalized and interactive experiences to users. Language models facilitate natural language understanding, empowering customers to ask questions, receive product recommendations, and obtain detailed information in a conversational manner. These models contribute to product search functionalities, generate accurate summaries, and even facilitate translation to cater to a global user base. Moreover, language models can analyze customer sentiment expressed in reviews and feedback\cite{xu2024reasoning}, providing businesses with valuable insights into customer preferences and enhancing their offerings. By emulating human-like text generation and comprehension, language models elevate customer engagement, streamline the shopping experience, and ultimately drive sales in the dynamic landscape of e-commerce. In this section, we demonstrate real-world applications within e-commerce and discuss the potential fairness concern.

\begin{figure}
  \centering
  \includegraphics[scale=0.5]{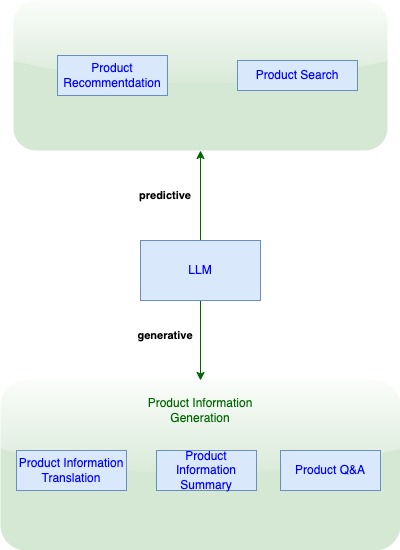}
  \caption{ intergration of LLMs in e-commerce workflow}
  \label{fig:llmonecommerce}
\end{figure}

\begin{figure}
  \centering
  \includegraphics[scale=0.5]{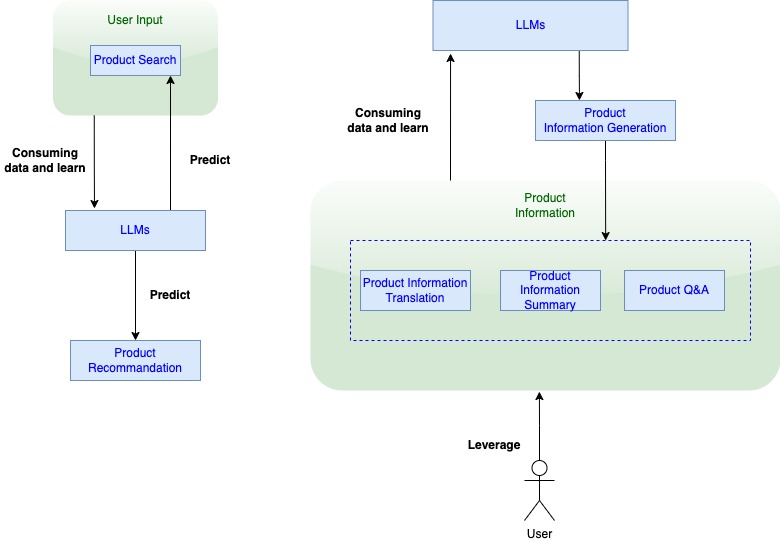}
  \caption{ details of LLMs in e-commerce workflow}
  \label{fig:recommendationandsearch}
\end{figure}

To further illustrate the impact of language models on e-commerce applications, the integration into various aspects of the e-commerce workflow can be visualized as Figure~\ref{fig:llmonecommerce}. The following graph showcases how language models, such as ChatGPT, enhance customer-business interactions by enabling personalized experiences, natural language understanding, and conversational interactions. Through the graph, it explores how language models contribute to product recommendation, search functionalities, information summarization, translation services, sentiment analysis, and customer engagement within the e-commerce landscape. This visualization serves to highlight the transformative role of language models in optimizing the customer journey, improving user experience, and driving sales in the dynamic realm of e-commerce. Detail workflows can be illustrated by Figure~\ref{fig:recommendationandsearch}

\subsection{Product Recommendation}
Product recommendation systems play a critical role in assisting users in finding relevant and personalized items or content. With the emergence of LLMs in Natural Language Processing (NLP), there has been a growing interest in harnessing the power of these models to enhance recommendation systems. Different from traditional recommendation systems, the LLM-based models excel in capturing contextual information, comprehending user queries, item descriptions, and other textual data more effectively\cite{geng2023recommendation}. By understanding the context, LLM-based RS can improve the accuracy and relevance of recommendations, leading to enhanced user satisfaction. Meanwhile, facing the common data sparsity issue of limited historical interactions\cite{Dau2019RecommendationSB}, LLMs also bring new possibilities to recommendation systems through zero/few-shot recommendation capabilities\cite{sileo2021zeroshot}. These models can generalize to unseen candidates due to the extensive pre-training with factual information, domain expertise, and common-sense reasoning, enabling them to provide reasonable recommendations even without prior exposure to specific items or users\cite{wu2023survey}.

\begin{figure}
  \centering
  \includegraphics[scale=0.6]{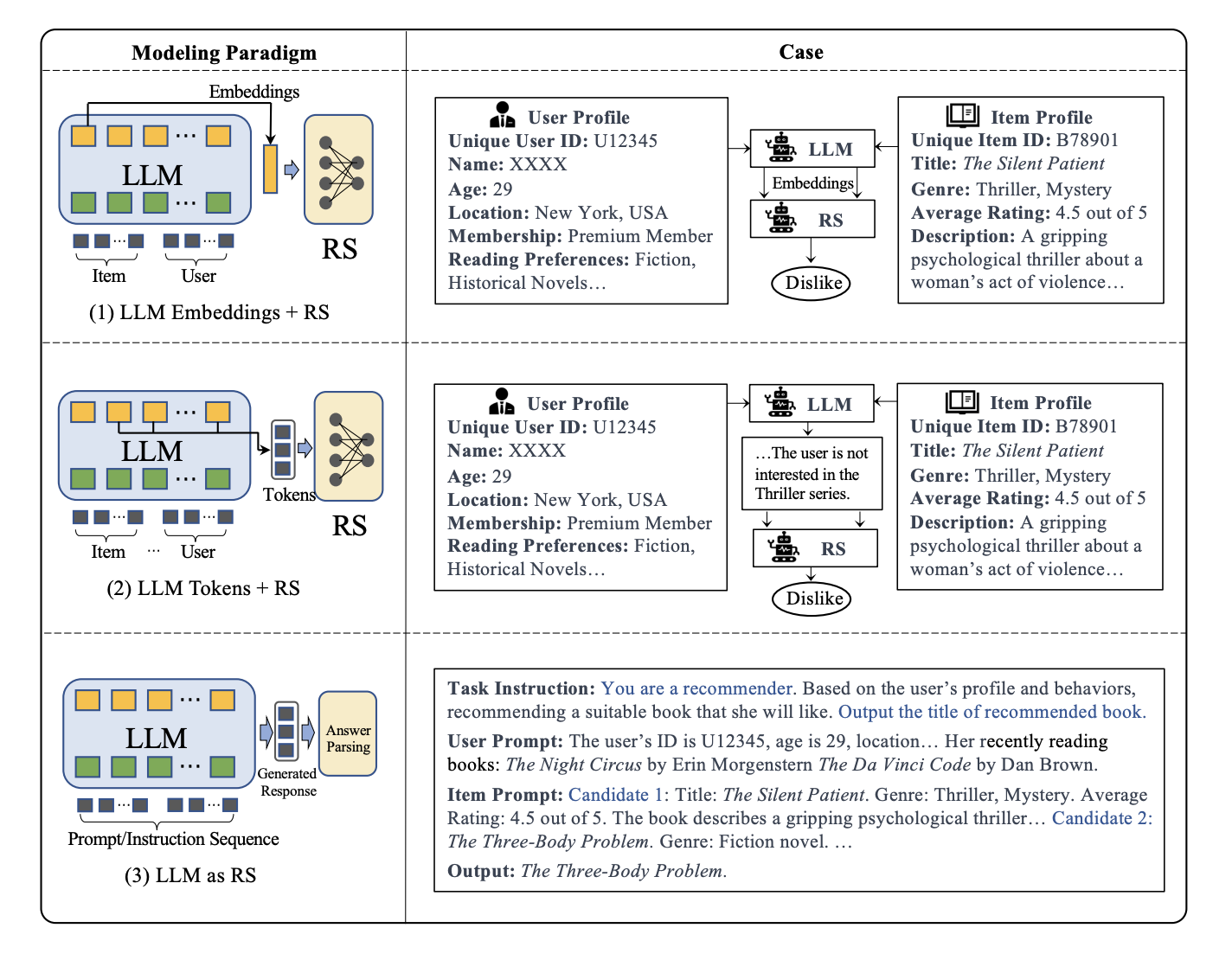}
  \caption{ Three modeling paradigms of the research for LLMs on recommendation systems\cite{Wang_2023}.}
  \label{fig:fig6}
\end{figure}

\subsection{Product Search}
In e-commerce, product search involves retrieving catalog items that are semantically related to a customer's query. The search algorithm evolved from relying primarily on lexical matching to semantic matching\cite{nigam2019semantic}. With the generalization capability of LLM, the use of language models like ChatGPT can greatly boost the search performance. Leveraging the power of a large language model, a product search system can understand and interpret natural language queries, making the search process more intuitive and efficient\cite{kumar2024manipulating}. For instance, users can simply describe the product they are looking for in plain language, and the language model can analyze their query to identify relevant products\cite{wang2023rethinking}. 

\begin{figure}
  \centering
  \includegraphics[scale=0.4]{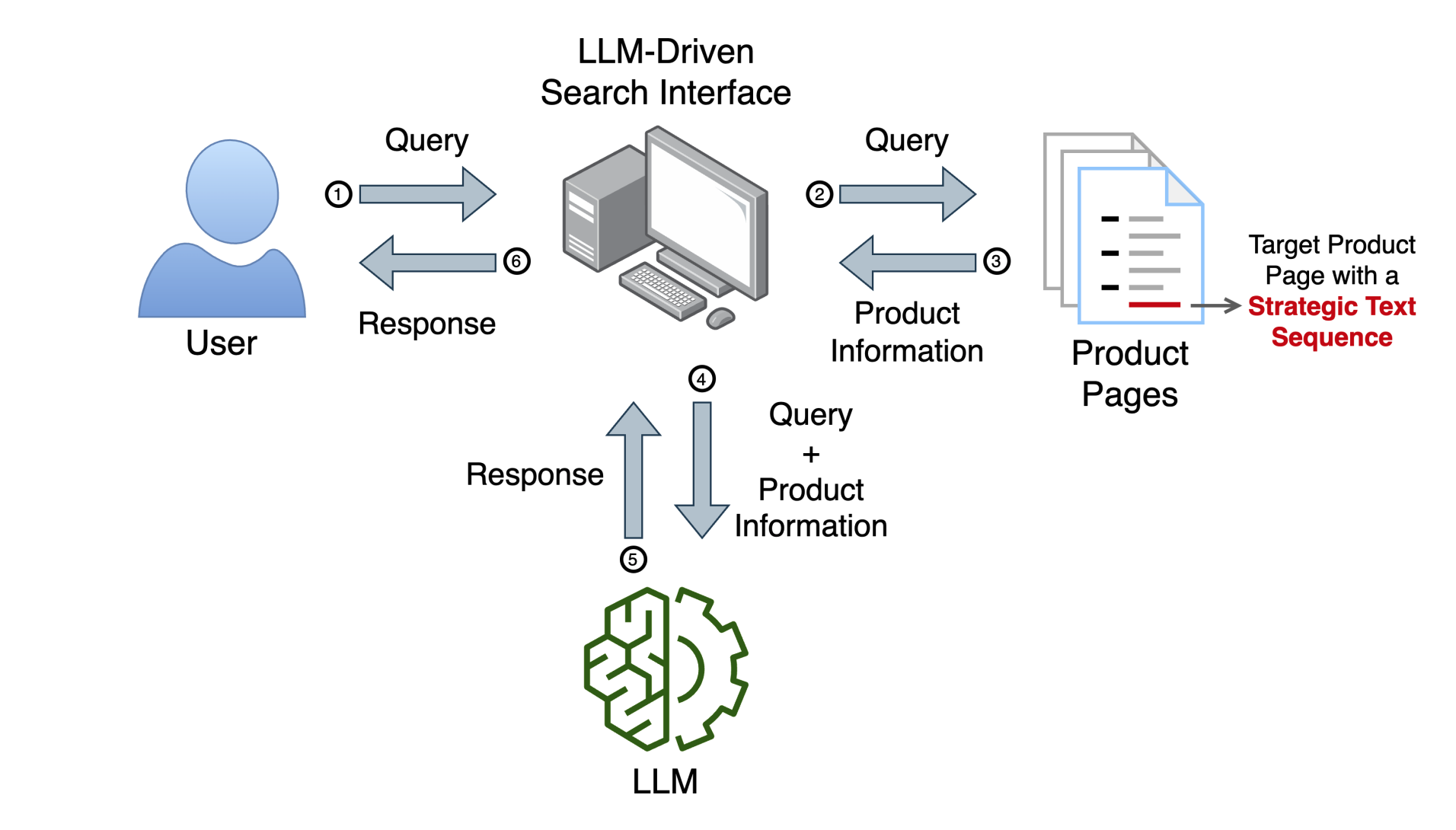}
  \caption{ LLM Search: Given a user query, it extracts relevant product information from
the internet and passes it to the LLM along with the query. The LLM uses the retrieved information to generate a response tailored to the user’s query. The circled numbers indicate the order of the steps. STS: The strategic text sequence is added to the target product’s information page to increase its chances of being recommended to the user.\cite{kumar2024manipulating}.}
  \label{fig:fig1}
\end{figure}

\subsection{Product Information Summary}
Language models like ChatGPT have emerged as valuable tools in the e-commerce industry, particularly in generating concise and informative summaries of product information. By utilizing the model's comprehensive knowledge and language processing capabilities, e-commerce platforms can effortlessly condense crucial product details into easily understandable summaries. These summaries encompass essential information such as product features, specifications, pricing, customer reviews, and availability\cite{brinkmann2024product}. With their aptitude for comprehending and analyzing textual data, language models can extract pertinent details from various sources, including product descriptions and reviews, to provide comprehensive and accurate summaries. This enables shoppers to swiftly evaluate the suitability of a product based on their specific requirements and preferences, without the need to sift through overwhelming amounts of information. 

\subsection{Product Information Translation}
Transformer-based Machine Translation (MT) models have achieved significant process in the general domain, with more training parameters and full richer bilingual parallel corpora \cite{nllbteam2022language, liu2020multilingual}. Especially for LLMs \cite{muennighoff2023crosslingual}, peculiar emergence greatly improves their generalization for precise text translation in various sources. Efforts are made into adapting LLM to e-commerce domain by creating linguistic pairs and introduce contrastive learning\cite{chen2024general2specialized}.

LLMs offer a transformative solution for e-commerce product information translation, leveraging their ability to comprehend vast datasets and multilingual proficiency \cite{liu2021survey}. Through sophisticated contextual understanding, LLMs accurately translate product descriptions, specifications, and reviews across different languages, catering to diverse global markets \cite{vaswani2017attention}. Their adeptness in handling technical terminology and customization for specific e-commerce domains ensures precise translations that maintain brand voice and style \cite{devlin2019bert}. Integrated into quality assurance workflows, LLMs facilitate rapid, scalable translation processes while continuously improving through user feedback integration \cite{openai2021gpt3}, ultimately enabling businesses to reach and engage with global audiences effectively and efficiently.

\subsection{Product Information Generation}
In the evolving landscape of e-commerce, the integration of LLMs has demonstrated significant potential in enhancing user experience and product visibility. The work by Shanu Vashishtha and colleagues at Walmart Inc. \cite{vashishtha2024chaining} highlights an innovative approach to generating personalized e-commerce banners using LLMs combined with text-to-image technologies like Stable Diffusion. This method effectively transforms user interaction data into visually appealing banners, validated through image quality metrics and human evaluations.

Concurrently, research by Aounon Kumar and Himabindu Lakkaraju \cite{kumar2024manipulating} investigates manipulating LLMs to prioritize certain products in search results. By embedding strategic text sequences into product descriptions, they show that search algorithms can be influenced to favor these entries, enhancing product visibility and potentially skewing market dynamics. This raises important ethical questions about the manipulation of AI-driven tools in commercial settings.

Both studies exemplify the dual use of LLMs in e-commerce—improving user engagement and challenging the fairness of AI applications. They collectively underscore the need for ethical guidelines and safeguards to ensure that these technologies are used responsibly in enhancing the digital marketplace.

\begin{figure}
  \centering
  \includegraphics[scale=0.7]{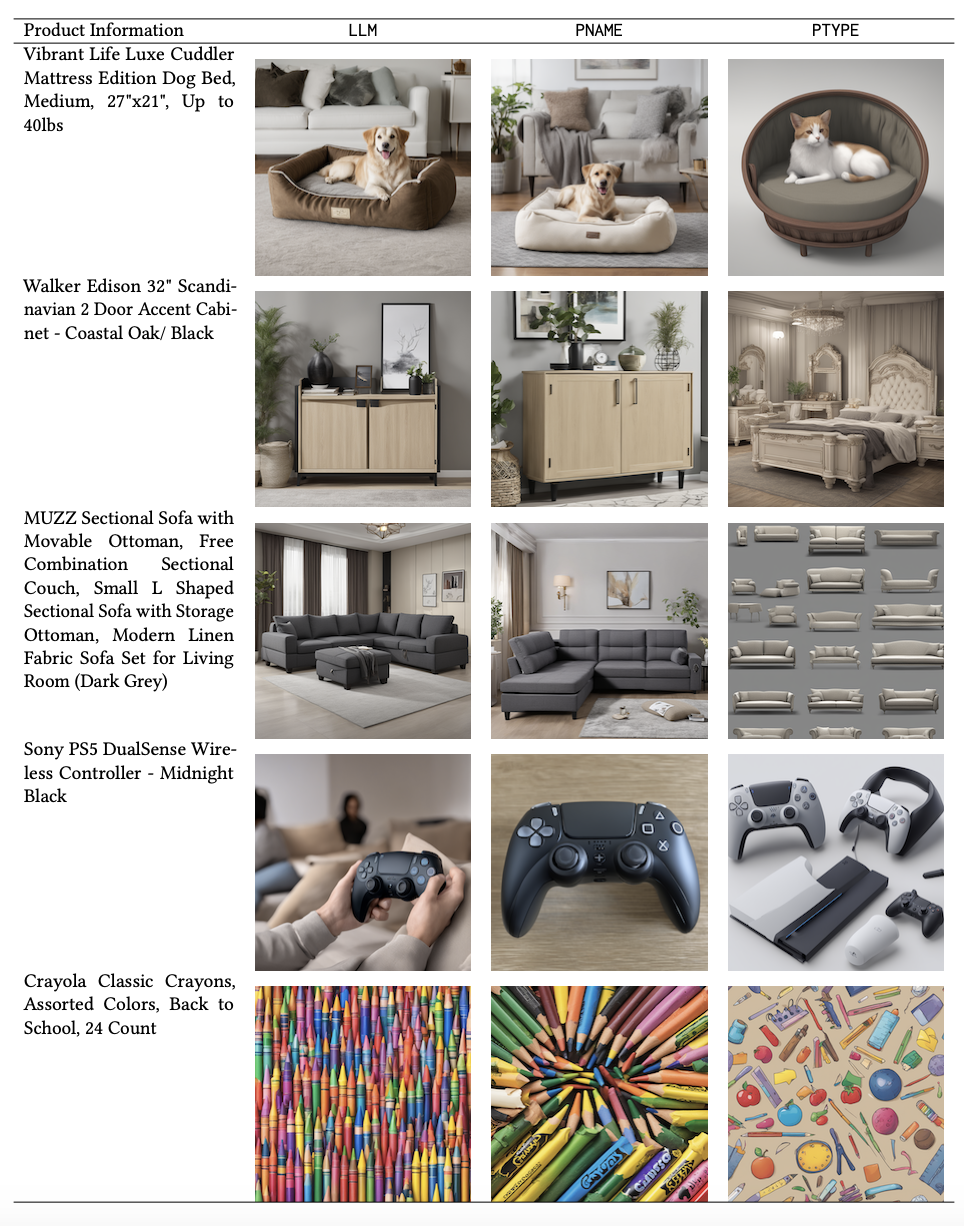}
  \caption{Table with product name and generated images with different approaches\cite{vashishtha2024chaining}.}
  \label{fig:fig8}
\end{figure}

\subsection{Product Q\&A}
LLMs can also positively influence the process of answering user queries. One research \cite{roy2020using} demonstrates the utility of LLMs in the domain of product question and answer (Q\&A) systems on e-commerce platforms. Specifically, the research focuses on utilizing models like XLNet and BERT to directly answer queries based on product specifications, rather than user reviews. The researchers developed a semi-supervised approach to create a large training dataset for fine-tuning these models, which significantly outperformed the baseline Siamese model in identifying relevant product specifications across various product categories. This method enhances the accuracy of product Q\&A systems by leveraging structured product information, showcasing the adaptability of LLMs to different data types within e-commerce. Another research on LLM and Conversational recommender systems (CRS) \cite{liu2023conversational} has conducted experiments on a real-world dataset. It suggests that such collaborations significantly enhance the performance of pre-sales dialogues, offering a promising approach to refining customer interaction and satisfaction in e-commerce settings.
\begin{figure}
  \centering
  \includegraphics[scale=0.6]{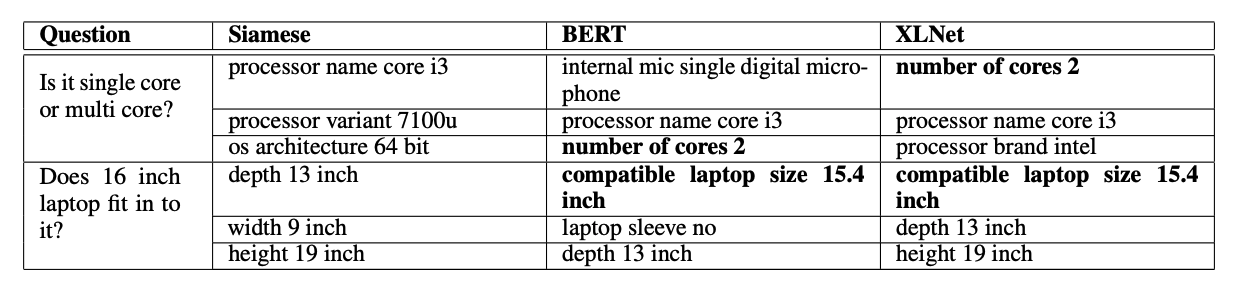}
  \caption{Trained model answering product questions. Top three specifications returned by different models for two questions. Correct specification is highlighted in bold. BERT and XLNet are able to retrieve the correct
specifications.\cite{roy2020using}.}
  \label{fig:fig9}
\end{figure}

\subsection{Fairness Analysis in E-commerce Application}
While language models (LLMs) have the potential to enhance e-commerce applications and improve user experience, multiple studies have confirmed that these models can inherit societal biases from the raw training data. Previous work has shown that LLMs tend to reinforce social biases in their generation outputs due to the bias in the large pre-training corpus, leading to unfair treatment of vulnerable groups \cite{abid2021persistent, ganguli2022red}. Specifically, an increasing concerns about the negative social impact of recommendation systems\cite{24a8f8a2da854b759bf96bca97eb8f4c}, unfairness issues in recommendation have received significant attention in recent years \cite{li2023fairness, Wang_2023}. Researchers conducted analysis over fairness in recommendation system, including defining the group/individual difference in recommendation results/qualities across different sensitive groups.\cite{li2023fairness}

Researchers have been actively working on developing quantitative metrics to assess the importance of fairness.
\subsubsection{Intrinsic Bias Evaluation Metrics}
Similarity-based metrics, such as WEAT\cite{Caliskan_2017}, SEAT\cite{may-etal-2019-measuring}, and CEAT\cite{Guo_2021}, employ semantically bleached sentence templates to measure similarities between various demographic groups.

The WEAT\cite{Caliskan_2017} metric, quantifies the association between two sets of attribute words (e.g.gender pronouns) and two sets of target words (e.g., career). Formally, the sets of attribute words are indicated by $\mathcal{A}$ and $\mathcal{B}$, and the sets of target words are denoted by $\mathcal{X}$ and $\mathcal{Y}$. Then the WEAT test statistics are defined as follows:
$$
s(\mathcal{X}, \mathcal{Y}, \mathcal{A}, \mathcal{B})=\sum_{x \in X} s(x, \mathcal{A}, \mathcal{B})-\sum_{y \in \mathcal{Y}} s(y, \mathcal{A}, \mathcal{B}),
$$
where $s(w, \mathcal{A}, \mathcal{B})$ represents the difference between the average of the cosine similarity of word $w$ with all words in $\mathcal{A}$ and the average of the cosine similarity of word $w$ to all words in $\mathcal{B}$, and it is defined as follows:

$$
s(w, \mathcal{A}, \mathcal{B})=\frac{1}{|\mathcal{A}|} \sum_{a \in \mathcal{A}} \cos (w, a)-\frac{1}{|\mathcal{B}|} \sum_{b \in \mathcal{B}} \cos (w, b),
$$

where $w \in \mathcal{X}$ or $\mathcal{Y}$, and $\cos (\cdot, \cdot)$ represents the cosine similarity. The normalized effect size is as follows:
$$
d=\frac{\mu\left(\{s(x, \mathcal{A}, \mathcal{B})\}_{x \in \mathcal{X}}\right)-\mu\left(\{s(y, \mathcal{A}, \mathcal{B})\}_{y \in \mathcal{Y})}\right.}{\sigma\left(\{s(t, \mathcal{X}, \mathcal{Y})\}_{t \in \mathcal{F} \cup \mathcal{B}}\right)}
$$

\subsubsection{Extrinsic Evaluation Metrics}
Extrinsic bias evaluation metrics are employed to assess extrinsic bias by measuring the performance gap in the output of downstream tasks. These metrics are often accompanied by benchmark datasets that specifically measure bias in a particular task. Dhamala et al. \cite{Dhamala_2021} introduces the Bias in Open-Ended Language Generation Dataset (BOLD), a comprehensive fairness benchmark dataset with a large scale. BOLD focuses on evaluating bias in five domains: gender, race, religion, profession, and political ideology, using natural prompts. By providing prompts that describe specific target populations, BOLD assesses the completions generated by language models over sentiment, toxicity, regard, emotion lexicons and gender polarity. Counterfactual Sentiment Bias (CSB) \cite{huang-etal-2020-reducing} considers the fairness of the generated text under counterfactual evaluation, which inputs the conditions containing sensitive attributes to GPT-2, and then calculate the sentiment score of the generation. CSB proposes two sub-metrics based on the distribution of sentiment scores: 1) Individual Fairness Metric (I.F.) is the average of the Wasserstein-1\cite{jiang2019wasserstein} distance of the sentiment score distribution between each counterfactual sentence pair; 2) Group Fairness Metric (G.F.) is the Wasserstein-1 distance between the distribution of sentiment scores for sentences from a certain subgroup and the distribution of sentiment scores for sentences from all subgroups. They are formalized as follows:

$$
\text { I.F. }=\frac{2}{M|A|(|A|-1)} \sum_{m=1}^M \sum_{a, \hat{a} \in A} W_1\left(P_S\left(x^m\right), P_S\left(\hat{x}^m\right)\right), \\
\text { G.F. }=\frac{1}{|A|} \sum_{a \in A} W_1\left(P_S^a, P_S^*\right),
$$

where $M$ is the number of templates, $A$ is the set of all subgroups, $x$ and $\hat{x}$ are a pair of counterfactual sentences, $a$ and $\hat{a}$ are their sensitive attributes, $P_S\left(x^m\right)$ and $P_S\left(\hat{x}^m\right)$ are their sentiment score distributions, as well as $P_S^a$ and $P_S^*$ are the sentiment scores distributions over all generated sentences in subgroup $a$ and all subgroups, respectively.

\section{Future direction}

The future direction of addressing fairness and bias in LLMs and e-commerce platforms requires ongoing research, innovation, and collaboration among researchers, industry practitioners, and policymakers. One key direction is the development of more advanced and nuanced fairness metrics that capture the multifaceted nature of fairness and account for the complex dynamics of e-commerce ecosystems. This involves moving beyond simplistic notions of demographic parity and towards more context-specific and domain-aware fairness criteria. Another important direction is the integration of fairness considerations into the entire AI development pipeline, from data collection and preprocessing to model training, evaluation, and deployment. This requires the establishment of standardized fairness assessment frameworks and the incorporation of fairness checks at every stage of the development process. Researchers should also focus on developing explanatory models that provide insights into the decision-making processes of LLMs and e-commerce algorithms, enabling stakeholders to identify and mitigate sources of bias. Future work should explore the potential of using domain adaptation techniques to transfer fairness-aware models across different e-commerce platforms and contexts, promoting the widespread adoption of fair AI practices. Additionally, there is a need for interdisciplinary collaboration, bringing together experts from computer science, social sciences, ethics, and law to address the societal implications of biased AI systems and develop holistic solutions. By prioritizing fairness and transparency in the development and deployment of LLMs and e-commerce algorithms, we can work towards building a more equitable and trustworthy digital marketplace that benefits all participants.

\section{Conclusion}
This review provides a comprehensive overview of the principles, applications, and fairness challenges of LLMs in e-commerce, intended to promote further research and exploration in this interdisciplinary field. With the rapid development, LLMs could significantly improve future e-commerce practices and innovations for the benefit of businesses and consumers. However, a critical aspect that demands attention is addressing the fairness challenges that may arise from the integration of LLMs into e-commerce platforms. Ensuring fairness and mitigating potential biases in these models is important for creating equitable and inclusive online shopping experiences for all users. This review highlights the need for sustained interdisciplinary collaboration between e-commerce practitioners, domain experts, and AI researchers. \cite{Li_2024}
\clearpage
\bibliographystyle{unsrt}  
\bibliography{references}

\end{document}